\documentclass[acmsmall]{acmart}
\AtBeginDocument{%
  }

\setcopyright{none}

\usepackage{graphicx}
\usepackage{amsmath}
\usepackage{amssymb}
\usepackage{hyperref}
\usepackage{tikz}
\usepackage{booktabs}
\usepackage{caption}
\usepackage{array}
\usepackage{multirow}
\usepackage{amssymb} 
\usepackage{pifont}  
\usepackage{algorithm}
\usepackage{algorithmic}
\usetikzlibrary{shapes.geometric, arrows, positioning}
\tikzstyle{process} = [rectangle, minimum width=2.5cm, minimum height=1.5cm, text centered, draw=black, fill=blue!20]
\tikzstyle{data} = [rectangle, minimum width=2cm, minimum height=1.2cm, text centered, draw=black, fill=green!20]
\tikzstyle{arrow} = [thick,->,>=stealth]

\begin{document}


\title{Neuro-Symbolic Manipulation Understanding with Enriched Semantic Event Chains}

\author{Fatemeh Ziaeetabar}
\authornote{Corresponding author}
\email{fziaeetabar@ut.ac.ir}
\orcid{0000-0003-1159-3588}

\affiliation{%
  \institution{Department of Computer Science, School of Mathematics, Statistics and Computer Science, College of Science, University of Tehran}
  \city{Tehran}
  \country{Iran}
}

%


\begin{abstract}
Robotic systems operating in human environments must reason about how object interactions evolve over time, which actions are currently being performed, and what manipulation step is likely to follow.
Classical enriched Semantic Event Chains (eSECs) provide an interpretable relational description of manipulation, but remain primarily descriptive and do not directly support uncertainty-aware decision making.
In this paper, we propose \textit{eSEC--LAM}, a neuro--symbolic framework that transforms eSECs into an explicit event-level symbolic state for manipulation understanding.
The proposed formulation augments classical eSECs with confidence-aware predicates, functional object roles, affordance priors, primitive-level abstraction, and saliency-guided explanation cues.
These enriched symbolic states are derived from a foundation-model-based perception front-end through deterministic predicate extraction, while current-action inference and next-primitive prediction are performed using lightweight symbolic reasoning over primitive pre-- and post--conditions.
We evaluate the proposed framework on EPIC-KITCHENS-100, EPIC-KITCHENS VISOR, and Assembly101 across action recognition, next-primitive prediction, robustness to perception noise, and explanation consistency.
Experimental results show that eSEC--LAM achieves competitive action recognition, substantially improves next-primitive prediction, remains more robust under degraded perceptual conditions than both classical symbolic and end-to-end video baselines, and provides temporally consistent explanation traces grounded in explicit relational evidence.
These findings demonstrate that enriched Semantic Event Chains can serve not only as interpretable descriptors of manipulation, but also as effective internal states for neuro--symbolic action reasoning.
\end{abstract}
\maketitle

\section{Introduction}
\label{sec:introduction}
Understanding manipulation actions from visual observations is a fundamental requirement for robotic systems operating in human-centered environments.
In practical settings such as human--robot collaboration, assistive robotics, and learning from demonstration, a robot must do more than classify an observed action.
It must interpret the evolving interaction between hands, tools, and manipulated objects, anticipate plausible next actions or intentions, and provide transparent justifications for its decisions~\cite{argall2009lfd,adama2018assistive,laplaza2025contextual,sridharan2019explanations}.

These requirements become particularly challenging in realistic scenes, where manipulation unfolds under clutter, occlusion, appearance variation, and temporally changing object interactions and roles~\cite{kaelbling2015multiple,li2026egocentricsurvey,shiota2024hocl,zheng2025embodiedsurvey}.

Large-scale benchmarks such as EPIC-KITCHENS-100~\cite{Damen2022RESCALING} and Assembly101~\cite{Sener_2022_CVPR} illustrate this difficulty, as they contain fine-grained object-centered interactions embedded in long and diverse action sequences.

Recent progress in video understanding has been driven largely by end-to-end deep architectures that learn appearance and motion representations directly from raw visual input.
Representative examples include efficient temporal models such as TSM~\cite{lin2019tsm}, large-scale egocentric learning frameworks such as Ego-Exo~\cite{grauman2023egoexo}, structured graph-based models such as MS-G3D~\cite{Liu_2020_CVPR}, and more recent interaction-aware or transformer-based approaches such as HOCL~\cite{shiota2024hocl} and HandFormer~\cite{shamil2024handformer}.
While these methods achieve strong empirical performance, they typically entangle perception and reasoning within black-box feature spaces.
As a consequence, they offer limited interpretability, provide weak control over the relational structure of manipulation processes, and make it difficult to perform systematic error analysis or incorporate explicit prior knowledge about objects, roles, and action structure.
For robotics applications, where traceability and controllability are often as important as predictive accuracy, these limitations remain significant.

A complementary line of research has therefore focused on symbolic and relational representations for manipulation understanding.
Among these, Semantic Event Chains and their enriched variants provide an interpretable abstraction of contact and spatial transitions between interacting entities~\cite{ziaeetabar2017semantic, ziaeetabar2018prediction, ziaeetabar2018recognition}.
In our previous work, eSECs were shown to be effective as structured relational descriptors for manipulation analysis and limited action recognition/prediction.
However, these representations remained primarily descriptive.
They did not explicitly model perceptual uncertainty, lacked semantic abstractions such as functional object roles and affordance priors, and did not provide a principled mechanism for primitive-level decision making, short-horizon action reasoning, or explanation generation.
In other words, prior eSEC-based approaches represented manipulation well, but they did not yet function as an internal action model.

\begin{figure*}[t]
    \centering
    \includegraphics[width=\textwidth]{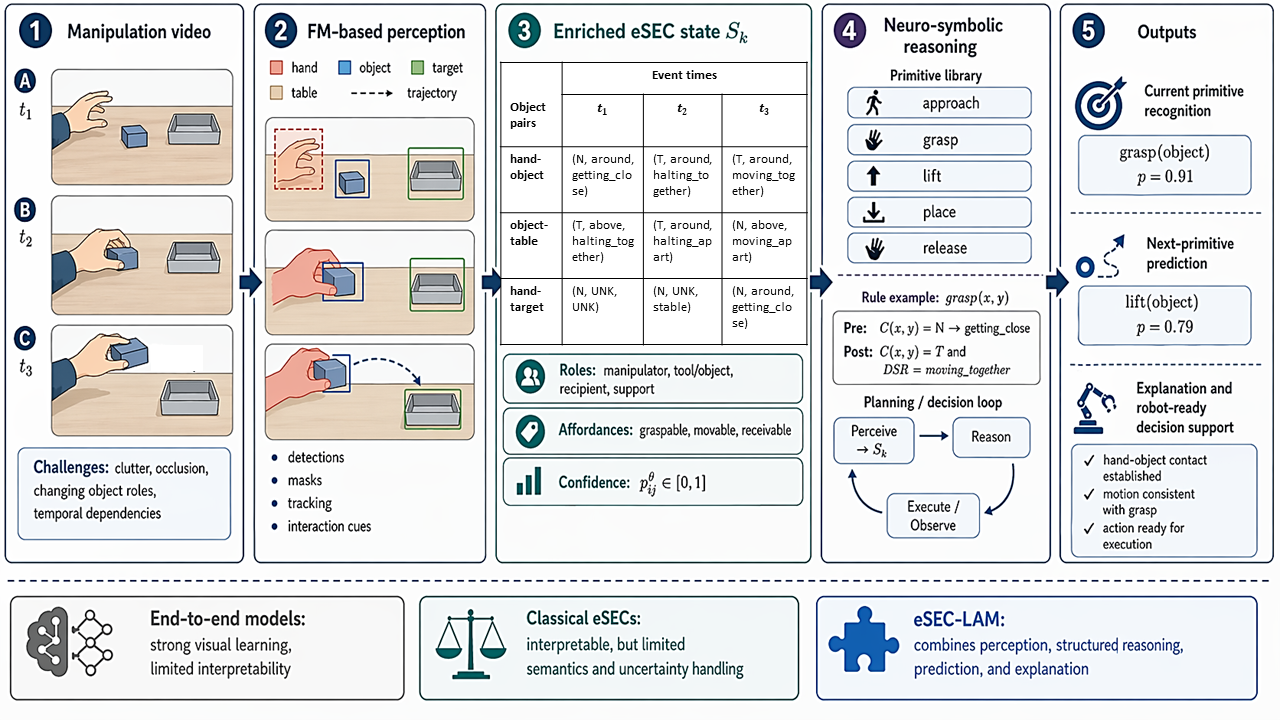}
    \caption{Motivation and overview of the proposed eSEC--LAM framework. Starting from manipulation video, a foundation-model-based perception module extracts object- and interaction-level cues, which are converted into symbolic predicates and organized into an enriched eSEC state. This state augments classical eSECs with confidence, affordances, roles, and primitive-level reasoning, enabling neuro-symbolic decision making for current-action inference, next-primitive prediction, and explanation generation. The bottom row conceptually contrasts the proposed framework with end-to-end visual models and classical eSEC representations.}
    \label{fig:intro_overview}
\end{figure*}

In this paper, we move beyond descriptive relational encoding and introduce a neuro--symbolic formulation for manipulation understanding.
Specifically, we transform enriched Semantic Event Chains into an explicit event-level symbolic state that supports action inference, next-primitive prediction, and explanation.
The central idea is to preserve the interpretability and compositional structure of eSECs while augmenting them with the semantic and uncertainty-aware elements required for decision making.
To this end, we incorporate confidence-aware predicates, functional object roles, affordance priors, primitive-level abstraction, and saliency-guided explanation cues into the eSEC representation.
These enriched symbolic states are derived from a foundation-model-based perception front-end, while action inference itself is performed through lightweight symbolic reasoning over primitive pre-- and post--conditions rather than through an end-to-end learned classifier.
The resulting framework, termed \textit{eSEC--LAM}, explicitly bridges visual perception, relational abstraction, and action-level reasoning.
Figure~\ref{fig:intro_overview} provides an intuitive overview of this formulation and highlights its conceptual position between end-to-end visual models and classical eSEC-based representations.

The proposed model enables several capabilities that are not jointly provided by previous eSEC-based approaches.
First, it performs ongoing action recognition from event-level symbolic states rather than raw frames.
Second, it supports next-primitive prediction by evaluating feasible future actions through structured symbolic constraints.
Third, it remains robust under perceptual degradation by incorporating confidence-aware relational abstraction.
Finally, it produces interpretable explanation traces grounded in the symbolic evidence that supports each decision.
Experiments on EPIC-KITCHENS-100~\cite{Damen2022RESCALING}, EPIC-KITCHENS VISOR~\cite{Darkhalil2022VISOR}, and Assembly101~\cite{Sener_2022_CVPR} show that the proposed framework achieves competitive action recognition, substantially improves next-primitive prediction, maintains stronger robustness to perception noise, and produces temporally consistent explanations grounded in explicit relational structure.

The main contributions of this paper are as follows:
\begin{itemize}
    \item We extend enriched Semantic Event Chains from an interpretable relational representation into a neuro--symbolic action model with an explicit event-level symbolic state for manipulation understanding.
    \item We introduce a set of principled enrichments---including confidence-aware predicates, object affordances, functional roles, primitive-level abstraction, and saliency-guided explanation cues---that make eSECs suitable for uncertainty-aware reasoning and prediction.
    \item We integrate foundation-model-based perception with deterministic predicate extraction and lightweight symbolic decision making, thereby combining robust visual front-ends with transparent relational reasoning.
    \item We demonstrate through quantitative and qualitative experiments that the proposed framework achieves competitive action recognition, substantially improved next-primitive prediction, stronger robustness under perceptual noise, and interpretable explanation generation across manipulation-centric video benchmarks.
\end{itemize}

The remainder of the paper is organized as follows.
Section~2 reviews the background and related foundations of the proposed approach.
Section~3 presents the proposed neuro--symbolic framework, including enriched state construction, perception-to-symbolic abstraction, decision making, and explanation generation.
Section~4 reports experimental results on action recognition, next-primitive prediction, robustness, ablation, and qualitative explanation consistency.
Finally, Section~5 concludes the paper and outlines future research directions.

\section{Related Work}

\subsection{Relational and Symbolic Representations for Manipulation Understanding}

A long-standing line of research in manipulation understanding has emphasized that actions can be represented more effectively through changes in object relations than through raw appearance alone.
Within this perspective, symbolic and relational representations offer an interpretable abstraction of manipulation processes by focusing on the evolving structure of contact, support, containment, and relative motion between manipulators and objects.
Our earlier work explored this direction through semantic spatial reasoning and Semantic Event Chain based formulations for manipulation analysis, showing that relational cues can support action recognition and prediction even when object-specific visual information is minimized~\cite{ziaeetabar2017semantic,ziaeetabar2018prediction,ziaeetabar2018recognition,ziaeetabar2020using}.

These studies demonstrated several important advantages of relational modeling.
First, they showed that event-centered symbolic abstraction can capture the core grammar-like structure of manipulation, beyond low-level trajectories or dense appearance features~\cite{worgotter2020humans}.
Second, they established that spatial and temporal relations provide a compact representation for action prediction, particularly in settings where semantic understanding must generalize across object instances and scene configurations~\cite{ziaeetabar2018prediction,ziaeetabar2020using}.
Third, later extensions toward graph-based and language-oriented manipulation understanding further confirmed that structured relational reasoning remains useful even when richer multimodal outputs such as multi-sentence video description are considered~\cite{ziaeetabar2024multi,Ziaeetabar2024}.

Despite these strengths, prior SEC/eSEC-based approaches remained primarily descriptive.
They were mainly used as interpretable representations for recognition and limited prediction, rather than as internal symbolic states for explicit decision making.
Moreover, they did not fully address perceptual uncertainty, dynamic functional role assignment, affordance-aware primitive reasoning, or saliency-grounded explanation.
The present work builds directly on this symbolic line, but moves from relational description to an uncertainty-aware neuro--symbolic action model that supports current-action inference, next-primitive prediction, and structured explanation within a unified framework.

\subsection{End-to-End Video Action Recognition and Interaction Modeling}

In parallel with symbolic approaches, recent progress in action understanding has been driven by end-to-end visual models that learn appearance and temporal dynamics directly from video.
Efficient temporal architectures such as TSM demonstrated that strong video understanding performance can be achieved with lightweight temporal feature shifting~\cite{lin2019tsm}.
Graph-based formulations such as MS-G3D showed the effectiveness of structured spatio-temporal modeling for action understanding from motion representations~\cite{Liu_2020_CVPR}.
More recent transformer-based and interaction-aware approaches have further improved the modeling of fine-grained hand--object dynamics, especially in egocentric and manipulation-centered settings~\cite{Kwon_2021_ICCV,shiota2024hocl,shamil2024handformer}.

This trend has been reinforced by large-scale benchmarks and training resources for egocentric and procedural understanding.
In particular, datasets and benchmarks such as EPIC-KITCHENS-100, Assembly101, and Ego-Exo style learning settings have pushed the field toward increasingly realistic, long-horizon, and interaction-rich scenarios~\cite{Damen2022RESCALING,Sener_2022_CVPR,grauman2023egoexo}.
Such developments have made modern end-to-end architectures highly competitive for action recognition and forecasting.
At the same time, they have also highlighted the difficulty of interpreting why a model predicts a certain action, which object relations were decisive, and how a predicted future action is grounded in an explicit representation of the scene.

Interaction-aware methods partly address this limitation by incorporating contact, hand pose, or object state cues into the recognition pipeline~\cite{Kwon_2021_ICCV,shiota2024hocl,shamil2024handformer}.
However, these methods still typically encode the relevant structure implicitly inside learned feature spaces.
As a result, they remain limited in their ability to expose symbolic preconditions, explicitly track object roles, or produce traceable reasoning over temporally evolving relational states.
Our work differs from this line not by replacing modern visual learning, but by using strong perception as a front-end and shifting the actual action reasoning process to an explicit event-level symbolic layer.

\subsection{Manipulation Understanding in Robotics, Anticipation, and Explanation}

For robotics applications, manipulation understanding is not only a recognition problem, but also a reasoning problem.
In practical settings such as learning from demonstration, assistive robotics, and human--robot collaboration, a robot must interpret ongoing interactions, anticipate plausible next actions, and provide transparent feedback about its decisions~\cite{argall2009lfd,adama2018assistive,laplaza2025contextual,sridharan2019explanations}.
This is especially important when robots operate in cluttered and partially observed environments, where uncertainty and object interaction structure strongly influence the success of manipulation and planning~\cite{kaelbling2015multiple}.

Recent surveys further emphasize that realistic egocentric and object-centric manipulation understanding remains challenging because of clutter, occlusion, changing object roles, and the need to reason over evolving interactions rather than isolated frames~\cite{li2026egocentricsurvey,zheng2025embodiedsurvey}.
These challenges motivate representations that remain interpretable while still benefiting from strong perceptual front-ends.
At the same time, the growing influence of foundation models in embodied perception and multimodal action understanding suggests that large pre-trained models can provide robust low-level cues, but they do not by themselves resolve the need for explicit structured reasoning over manipulation events~\cite{ziaeetabar2025leveraging}.

The present work is positioned at the intersection of these directions.
It retains the object-centric interpretability of symbolic manipulation representations, incorporates the robustness of modern perception modules, and introduces an explicit decision-oriented symbolic state that supports anticipation and explanation.
In this sense, the proposed framework is closer to robotics-oriented reasoning systems than to purely end-to-end action recognizers: it uses perception to instantiate symbolic predicates, and then performs lightweight confidence-aware reasoning over action primitives, object roles, and affordances.
This combination enables action recognition, next-primitive prediction, and explanation within a single neuro--symbolic formulation.

\section{Method}
\label{sec:method}

\subsection{Overview of the Proposed eSEC--LAM}

\begin{figure}[t]
    \centering
    \includegraphics[width=\linewidth]{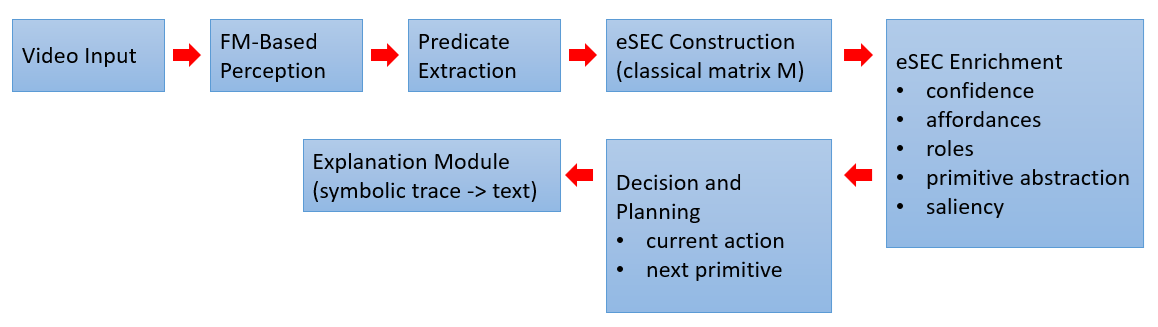}
    \caption{Technical overview of the proposed eSEC--LAM architecture.
A foundation--model perception module extracts object-, hand-, and interaction-level cues from video, which are converted into symbolic predicates and organised into an event--centric eSEC matrix $M$.
The eSEC is enriched with confidence scores, affordance priors, functional roles, macro--event abstractions, and saliency measures to form an explicit internal state.
A lightweight decision module operates on this symbolic state to infer the current manipulation primitive and predict plausible next primitives, and an explanation module returns the salient relational evidence underlying each inference step.}
    \label{fig:architecture}
\end{figure}

Having introduced the conceptual motivation of the proposed framework in Figure~\ref{fig:intro_overview}, we now describe its technical organisation.
Figure~\ref{fig:architecture} summarises the main computational modules of the proposed Large Action Model grounded in enriched Semantic Event Chains (eSEC--LAM).

Given an input video sequence, the system first processes visual observations through a perception module based on foundation models to extract high--level cues related to objects, hands, and their interactions.
These cues are subsequently mapped to symbolic predicates describing contact, static spatial, and dynamic relations between pairs of entities.
The resulting predicate streams are temporally abstracted into an event--centric eSEC matrix, which compactly represents relational state transitions over time.

While classical eSECs were originally introduced as descriptive representations for action recognition, we extend them here into an internal symbolic state suitable for reasoning.
To this end, we augment the eSEC formulation with confidence scores, functional role labels, object affordance priors, macro--event abstractions, and saliency measures.
Together, these extensions transform the eSEC from a passive relational descriptor into an actionable representation that can support inference, prediction, and explanation.

On top of the enriched eSEC state, a lightweight decision module evaluates relational patterns under confidence--aware pre-- and post--condition constraints to infer the currently ongoing manipulation primitive and to predict plausible next primitives.
The resulting decisions are grounded in the symbolic state and do not rely on end--to--end learned action classifiers.
Finally, an explanation module extracts a symbolic trace from the eSEC and the decision process, identifying the relational events that supported each inference step.
This trace can be directly inspected or rendered using simple, structured verbal templates.

This section proceeds in three stages.
First, we formalise the classical eSEC representation adopted from prior work~\cite{ziaeetabar2018recognition}.
Second, we introduce principled enrichments that equip eSECs with uncertainty awareness and semantic attributes required for reasoning.
Third, we describe how enriched eSECs interface with foundation--model perception and with the decision and explanation modules that together constitute the proposed neuro--symbolic Large Action Model.

\subsection{Classical Enriched Semantic Event Chains (eSECs)}

We first recall the classical formulation of Enriched Semantic Event Chains (eSECs), which constitute the representational backbone of the proposed model. 
We adopt the original event--centric definition introduced in prior work~\cite{ziaeetabar2017semantic}, while presenting a streamlined and implementation--oriented variant that facilitates integration with modern foundation--model perception pipelines.
This formulation preserves the core relational and temporal abstraction properties of eSECs, which are essential for manipulation understanding, while remaining agnostic to the specific visual front--end.

\subsubsection{Fundamental Entities and Pairwise Relations}

Let $O = \{o_1,\dots,o_N\}$ denote the set of fundamental entities participating in a manipulation episode, including manipulators (hands), manipulated objects, and supporting surfaces.
We define the set of ordered object pairs
\begin{equation}
    \mathcal{P} = \{(o_i,o_j) \mid o_i,o_j \in O,\ i \neq j\}.
\end{equation}
Time is discretised into frame indices $\tau \in \{1,\dots,T\}$.
Events occur at a subsequence of frames $\{t_1,\dots,t_K\} \subseteq \{1,\dots,T\}$ at which at least one relation between any pair changes.

For each pair $(o_i,o_j) \in \mathcal{P}$ and event index $k \in \{1,\dots,K\}$, we consider three types of relations:
\begin{itemize}
    \item a \emph{contact relation}
    \(
        r_{ij}^{\mathrm{C}}(t_k) \in \mathcal{R}_{\mathrm{C}} = \{T,N\},
    \)
    indicating whether the entities are in physical contact ($T$) or not ($N$);
    \item a \emph{static spatial relation}
    \(
        r_{ij}^{\mathrm{S}}(t_k) \in \mathcal{R}_{\mathrm{S}},
    \)
    describing coarse, time-invariant spatial configuration;
    \item a \emph{dynamic spatial relation}
    \(
        r_{ij}^{\mathrm{D}}(t_k) \in \mathcal{R}_{\mathrm{D}},
    \)
    characterising changes in relative position or motion patterns over time.
\end{itemize}

In accordance with prior eSEC formulations, we adopt the following relation sets:
\begin{equation}
\begin{aligned}
\mathcal{R}_{\mathrm{S}} &= 
\{\texttt{around}, \texttt{above}, \texttt{below}, \texttt{inside}\}, \\
\mathcal{R}_{\mathrm{D}} &= 
\{\texttt{getting\_close}, \texttt{moving\_apart}, \texttt{stable}, \\
&\quad \texttt{halting\_together}, \texttt{moving\_together}, 
\texttt{fixed\_moving\_together}\}.
\end{aligned}
\end{equation}
These relations capture the essential spatial and temporal interaction patterns required for manipulation understanding, while avoiding explicit reliance on object orientation or fine-grained kinematic estimation.
This choice ensures robustness under partial observability and compatibility with diverse perception backends.

\subsubsection{Event--Based Matrix Representation}

An enriched Semantic Event Chain is represented as a matrix $M$ whose rows correspond to ordered object pairs and whose columns correspond to discrete events.
Formally, the $k$--th column $c_k$ encodes all relations at event time $t_k$:
\begin{equation}
    c_k = \Big( \big(r_{ij}^{\mathrm{C}}(t_k), r_{ij}^{\mathrm{S}}(t_k), r_{ij}^{\mathrm{D}}(t_k)\big) \Big)_{(i,j) \in \mathcal{P}}.
\end{equation}
The full event--level eSEC for a sequence is then given by
\begin{equation}
    M = [c_1, c_2, \dots, c_K].
\end{equation}

New event times $t_k$ (and thus new columns $c_k$) are instantiated whenever at least one relation changes between consecutive frames:
\begin{equation}
    \exists (i,j)\in\mathcal{P},\ \exists \theta \in \{\mathrm{C},\mathrm{S},\mathrm{D}\}:
    \quad r_{ij}^{\theta}(t_k) \neq r_{ij}^{\theta}(t_{k-1}).
\end{equation}
This construction yields a compact temporal abstraction in which each column represents a meaningful relational transition rather than an individual video frame.

\subsubsection{Use in Prior Work and Limitations}

In previous works~\cite{ziaeetabar2017semantic, ziaeetabar2018prediction, ziaeetabar2018recognition, worgotter2020humans, ziaeetabar2020using, ziaeetabar2020spatio}, 
eSECs were primarily employed as interpretable relational descriptors for manipulation action recognition and limited action prediction.
Similarity measures over complete eSEC matrices and their prefixes were used to classify observed sequences and to infer likely continuations.
However, these approaches relied on hand--crafted RGB--D segmentation pipelines and treated eSECs mainly as passive descriptors, without explicitly supporting planning, compositional reasoning, or structured explanation.

In contrast, our goal is to retain the core event--centric relational structure of eSECs while:
(i) modernising the perceptual front--end through foundation models,
(ii) enriching the representation with additional semantics such as confidence, roles, and object affordances, and
(iii) transforming eSECs from a descriptive representation into an internal symbolic state suitable for decision making within a neuro--symbolic Large Action Model.
The following subsections detail these extensions.

\subsection{Extending eSECs for Neuro--Symbolic LAM}

While classical eSECs provide a compact and interpretable relational representation, they are not directly suitable as an internal state for planning, compositional reasoning, or structured explanation in Large Action Models. 
We therefore introduce a set of principled extensions that augment eSECs with uncertainty awareness, semantic object attributes, hierarchical abstractions, and decision saliency. 
These enrichments preserve the original event--centric formulation, while equipping the representation with the additional structure required for neuro--symbolic reasoning and action inference.

\begin{figure*}[t]
  \centering
  \includegraphics[width=\textwidth]{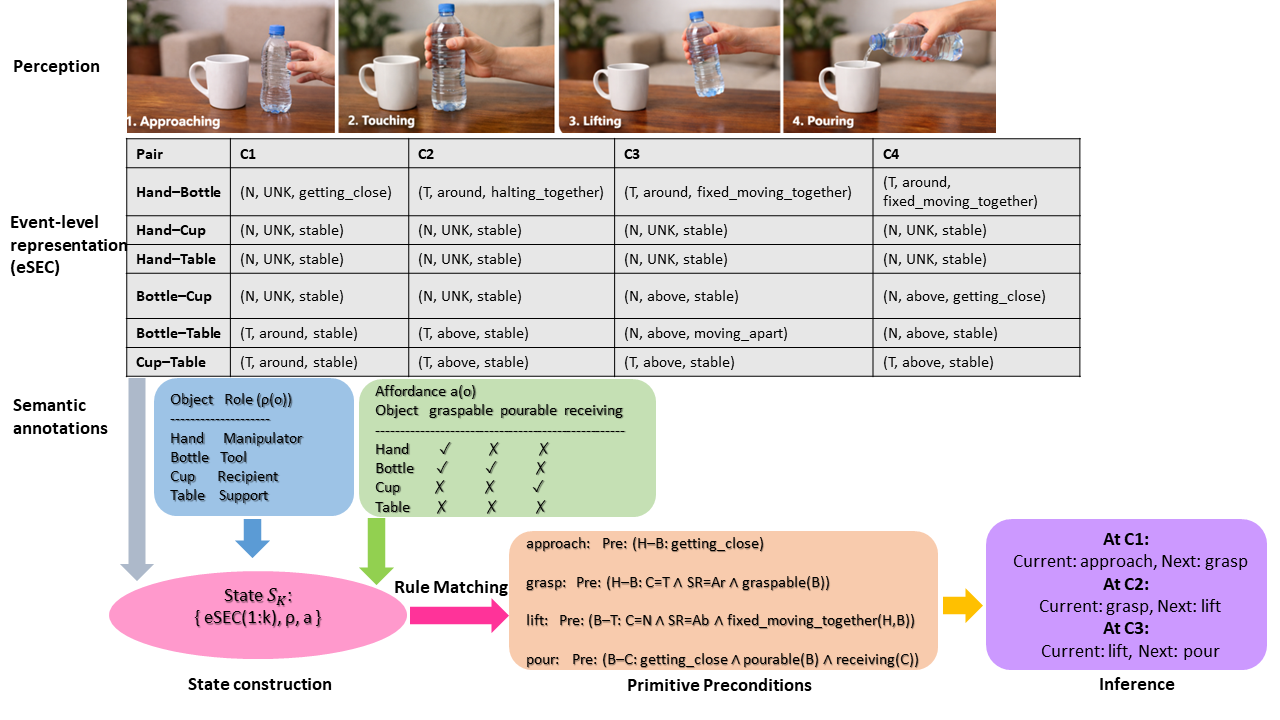}
  \caption{Extending event-level enriched Semantic Event Chains (eSECs) into a neuro--symbolic Large Action Model (LAM) for manipulation.
  FM-based perception extracts entities and pairwise relations, which are organized into an event-level eSEC using legal contact ($C$), static spatial relation ($SSR$), and dynamic spatial relation ($DSR$) labels.
  Object roles $\rho(o)$ and affordances $a(o)$ are attached as semantic annotations and aggregated into a symbolic state $S_k$ that integrates relational predicates ($c_k$), object roles $\rho(O)$, affordances $a(O)$, and confidence values $p_k$.
  Action primitives are inferred via rule matching over $S_k$ using a dictionary of primitive preconditions, enabling interpretable current-primitive inference and next-primitive prediction.}
  \label{fig:esec-lam}
\end{figure*}

Figure~\ref{fig:esec-lam} provides a concrete illustration of the proposed extensions on a representative manipulation sequence. 
Starting from FM-based perception, pairwise relations are organized into an event-level eSEC. 
Object-centric semantic annotations---roles $\rho(o)$ and affordances $a(o)$---are then attached and aggregated into a symbolic representation, which is later formalized as the state $S_k$.
Action primitives are subsequently inferred via rule matching against a dictionary of primitive preconditions.

\begin{figure}[t]
    \centering
    \includegraphics[width=\linewidth]{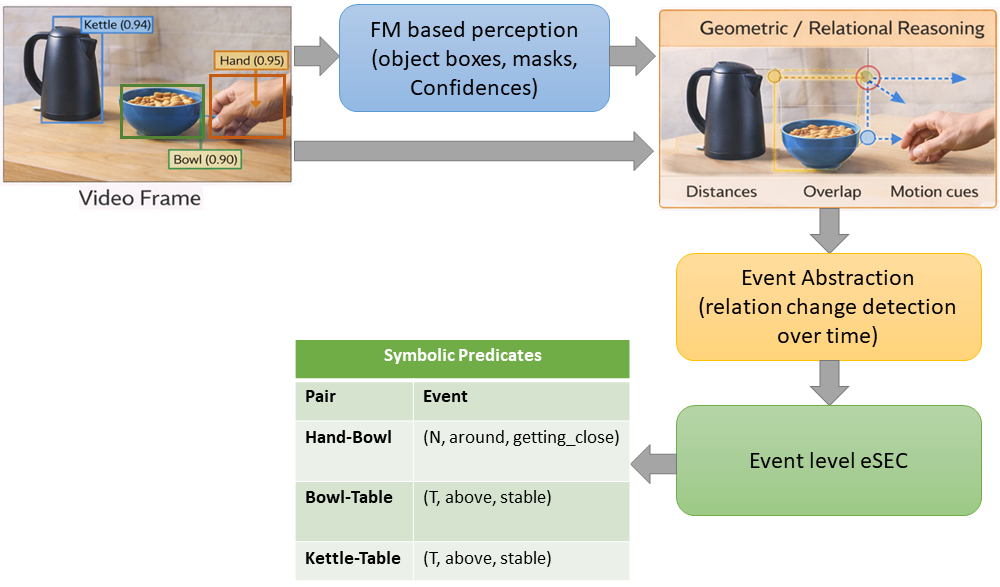}
    \caption{Overview of the perception-to-symbolic abstraction pipeline.
    Frame-level object detections produced by foundation-model-based perception are converted into geometric and relational cues, from which symbolic predicates are derived.
    Temporal abstraction over relation changes yields an event-level eSEC representation that serves as the symbolic state $S_k$ for subsequent neuro--symbolic reasoning.}
    \label{fig:perception-to-esec}
\end{figure}

Figure~\ref{fig:perception-to-esec} illustrates the perception-to-symbolic abstraction pipeline, showing how foundation-model-based visual outputs are converted into geometric relations and subsequently aggregated into an event-level eSEC, forming the symbolic state $S_k$ used by the neuro--symbolic Large Action Model.

\subsubsection{Confidence--Aware and Unknown Relations}

To account for perceptual uncertainty introduced by foundation--model extraction and subsequent geometric reasoning, we associate each symbolic relation with a confidence value.
For each object pair $(o_i,o_j) \in \mathcal{P}$, event index $k$, and relation type $\theta \in \{\mathrm{C},\mathrm{S},\mathrm{D}\}$, we define:
\begin{equation}
    r_{ij}^{\theta}(t_k) \in \mathcal{R}_{\theta} \cup \{\texttt{UNK}\}, 
    \qquad
    p_{ij}^{\theta}(t_k) \in [0,1].
\end{equation}
The label \texttt{UNK} denotes relations that are unobserved or insufficiently supported, for instance due to occlusions, detection failures, or ambiguous motion patterns.

Event boundaries are refined by incorporating confidence into the classical change criterion.
Specifically, a new event is instantiated at time $t_k$ only if
\begin{equation}
    \exists (i,j)\in\mathcal{P},\, \exists \theta\in\{\mathrm{C},\mathrm{S},\mathrm{D}\} :
    r_{ij}^{\theta}(t_k) \neq r_{ij}^{\theta}(t_{k-1}) 
    \;\wedge\; 
    \max_{\theta} p_{ij}^{\theta}(t_k) \ge \tau_{\mathrm{event}}.
\end{equation}
This refinement suppresses spurious event creation caused by noisy relation estimates while preserving the original event--centric abstraction.
Confidence values are subsequently used to formulate soft constraints in later reasoning stages, enabling decisions under partial observability without committing to hard thresholds.

\subsubsection{Object Affordance Augmentation}

To support primitive-level generalisation across object instances, we optionally associate each object with a set of semantic affordance attributes.
Let $\mathcal{A}$ denote a coarse affordance vocabulary, e.g.,
\begin{equation}
    \mathcal{A} = \{\texttt{graspable},\, \texttt{pourable},\, \texttt{receiving},\, \texttt{cuttable},\, \texttt{openable}\}.
\end{equation}
Each object $o \in O$ is assigned a (possibly multi-valued) affordance vector
\begin{equation}
    a(o) \in \{0,1\}^{|\mathcal{A}|},
\end{equation}
derived from object category information provided by the perception backbone or from lightweight domain priors.

Affordances are treated as weak semantic cues rather than strict constraints.
They remain constant throughout a sequence and may overlap (e.g., a cup may be both \texttt{pourable} and \texttt{receiving}).
In the proposed framework, affordances are used to bias primitive applicability and interpretation, rather than to enforce hard exclusion rules.
This design choice avoids brittle behaviour in ambiguous scenes and complements the more context-dependent notion of functional object roles introduced next.

\subsubsection{Functional Object Roles}

Beyond coarse affordance attributes, objects assume functional roles that depend on the interaction context within a manipulation episode.
We define a finite role set
\begin{equation}
    \mathcal{R}_{\mathrm{role}} 
    = \{\texttt{manipulator},\, \texttt{tool},\, \texttt{recipient},\, \texttt{support}\},
\end{equation}
and associate each object with a role assignment
\begin{equation}
    \rho: O \times t_k \rightarrow \mathcal{R}_{\mathrm{role}},
\end{equation}
which may change over time as the interaction evolves.

Roles are inferred from relational patterns captured in the enriched eSEC representation, rather than being fixed object properties.
For example, an object that is \texttt{holding}-related to a manipulator and exhibits \texttt{moving\_together} dynamics may assume the role of a \texttt{tool}, while another object that remains spatially stable and receives contents may assume the role of a \texttt{recipient}.
Supporting surfaces are identified through persistent \texttt{contact} and \texttt{stable} relations with other objects.

Functional roles constrain which object pairs are relevant for particular manipulation primitives and enable generalisation across object instances.
Crucially, role-based reasoning allows previously unseen objects to participate in inferred actions based on their interaction behaviour, even when their semantic category or affordance profile is ambiguous.

\subsubsection{Primitive--Level Abstraction of Event Chains}

While the event--centric eSEC matrix $M = [c_1,\dots,c_K]$ captures relational transitions at fine temporal resolution, higher--level manipulation understanding benefits from grouping such events into semantically meaningful action units, referred to as \emph{primitives}.
We therefore derive a higher--level abstraction
\[
P = (p_1,\dots,p_M), \quad M \le K,
\]
where each $p_m$ denotes a primitive inferred from one or more consecutive event columns in $M$.

Formally, we define a mapping
\begin{equation}
    \Phi: M \rightarrow P,
\end{equation}
which groups event columns into primitive segments based on characteristic relational patterns observed in manipulation sequences (e.g., approach, grasp, lift, pour, release).
This abstraction yields a discrete, temporally ordered action trajectory that serves as an intermediate representation between low-level relational events and higher-level reasoning components.
Symbolic pre--conditions and post--conditions are evaluated over this primitive abstraction in subsequent stages.

\subsubsection{Unified Symbolic State Representation}
\label{sec:state}

To enable structured reasoning and decision making, we consolidate the enriched eSEC representation and semantic annotations into a unified symbolic state.

At each event time $t_k$, we define the symbolic state as:
\begin{equation}
S_k := \{ c_k,\ \rho(O),\ a(O),\ p_k \},
\end{equation}
where:
\begin{itemize}
    \item $c_k$ denotes the $k$-th column of the event-level eSEC matrix, encoding all pairwise relational predicates $(r^C_{ij}, r^S_{ij}, r^D_{ij})$ at time $t_k$,
    \item $\rho(O)$ represents the set of functional role assignments for all objects,
    \item $a(O)$ denotes the object-level affordance attributes,
    \item $p_k$ collects the confidence values associated with all predicates in $c_k$.
\end{itemize}

This formulation provides a compact and explicit representation of the current relational configuration of the scene, enriched with semantic and uncertainty information.
Unlike classical eSEC formulations, which operate on the entire matrix $M$, the proposed state representation is event-local and supports incremental reasoning.

The state $S_k$ serves as the formal interface between perception and symbolic reasoning, and forms the basis for evaluating primitive preconditions, selecting action primitives, and predicting subsequent actions.

\subsection{FM-Based Perception and Predicate Extraction}
\label{sec:fm_perception}

To populate the enriched eSEC representation, we require symbolic predicates describing contact, static spatial, and dynamic relations between object pairs over time.
These predicates are derived from a perception pipeline built upon foundation models (FMs) for video understanding.
Rather than directly predicting symbolic relations in an end-to-end manner, the FM backbone provides mid-level visual cues—including object detections, segmentation masks, hand-localization signals, and interaction likelihoods—which are subsequently transformed into discrete relational predicates through explicit geometric and temporal reasoning.

This design separates perception from symbolic abstraction: foundation models supply robust, general-purpose visual evidence, while relational predicates are computed deterministically to ensure transparency and consistency with the classical eSEC formulation.
Each extracted predicate is coupled with a confidence score reflecting the reliability of the underlying perceptual evidence, in accordance with the confidence-aware extension introduced earlier.

\subsubsection{Perception Backbone and Low-Level Outputs}

Given an input frame $\tau$, the perception module produces a set of detected entities
\(
O(\tau) = \{o_1,\dots,o_{N(\tau)}\},
\)
including hands, manipulated objects, and supporting surfaces.
For each detected entity $o_i$, the module provides a bounding box $B_i(\tau)$, an optional segmentation mask $M_i(\tau)$, and a detection confidence score $s_i(\tau) \in [0,1]$.
For hands, the foundation model additionally outputs articulated keypoints describing hand pose.

Temporal tracking associates detections across frames, yielding object trajectories over time.
These trajectories form the basis for estimating relative motion patterns and interaction cues, such as approach, separation, and sustained contact.

All symbolic relations used in the eSEC representation are derived from these mid-level perceptual outputs.
Importantly, foundation models are not used to directly predict symbolic relations; instead, relations are computed through deterministic geometric and temporal rules applied to bounding boxes, masks, keypoints, and trajectories.
This separation preserves interpretability and maintains consistency with the classical eSEC formulation.

\subsubsection{Static Spatial Relation Estimation}

Let $(o_i,o_j) \in \mathcal{P}$ denote an ordered object pair observed at frame $\tau$.
Static spatial relations encode the instantaneous qualitative geometric configuration between entities and are evaluated independently at each frame.
For each pair, a static spatial relation $r_{ij}^{\mathrm{S}}(\tau)$ is deterministically estimated based on geometric criteria applied to bounding boxes $B_i(\tau)$, $B_j(\tau)$ and, when available, segmentation masks $M_i(\tau)$, $M_j(\tau)$.

We define the static spatial relation vocabulary as
\begin{equation}
\mathcal{R}_{\mathrm{S}} =
\{\texttt{inside},\ \texttt{on},\ \texttt{above},\ \texttt{below},\ \texttt{around}\},
\end{equation}
with an additional fallback label $\texttt{UNK}$ indicating insufficient evidence for any relation in $\mathcal{R}_{\mathrm{S}}$.

\paragraph{Inside.}
We assign $\texttt{inside}(i,j)$ if
\begin{equation}
    \frac{|M_i(\tau) \cap M_j(\tau)|}{|M_i(\tau)|} \ge \tau_{\mathrm{inside}},
\end{equation}
indicating that object $o_i$ is largely contained within object $o_j$.
When segmentation masks are unavailable, this relation is approximated using bounding-box inclusion tests, requiring $B_i(\tau)$ to be spatially contained within $B_j(\tau)$ up to a tolerance margin.

\paragraph{On.}
We assign $\texttt{on}(i,j)$ if (i) the horizontal projections of $B_i(\tau)$ and $B_j(\tau)$ exhibit sufficient overlap and (ii) the lower boundary of $B_i(\tau)$ lies within a small vertical tolerance of the upper boundary of $B_j(\tau)$, indicating support contact.

\paragraph{Above / Below.}
Let $c_i(\tau)=(x_i(\tau),y_i(\tau))$ and $c_j(\tau)=(x_j(\tau),y_j(\tau))$ denote the centroids of $B_i(\tau)$ and $B_j(\tau)$, respectively.
Assuming an image coordinate system with downward-increasing vertical axis, we assign
$\texttt{above}(i,j)$ if $y_i(\tau) + \delta_y \le y_j(\tau)$, and
$\texttt{below}(i,j)$ if $y_i(\tau) \ge y_j(\tau) + \delta_y$.

\paragraph{Around / Not-Around.}
The relation $\texttt{around}(i,j)$ captures a coarse notion of spatial proximity without relying on metric distance thresholds.
It is assigned when the spatial extents of $B_i(\tau)$ and $B_j(\tau)$ exhibit partial overlap or close adjacency, and neither containment nor support relations apply.
This relation represents configurations in which objects are spatially close or partially surrounding each other (e.g., during grasping or encircling motions), without committing to fine-grained distance quantization.

If the geometric configuration does not satisfy the criteria for \texttt{around} and no other relation in $\mathcal{R}_{\mathrm{S}}$ holds, no static spatial relation is assigned from $\mathcal{R}_{\mathrm{S}}$, and the pair is represented as $\texttt{UNK}$.

\paragraph{Relation Selection.}
When multiple conditions are simultaneously satisfied, a deterministic priority ordering is applied:
\begin{equation}
\texttt{inside} \succ \texttt{on} \succ \{\texttt{above},\texttt{below}\} \succ \texttt{around}.
\end{equation}
If none of the relations in $\mathcal{R}_{\mathrm{S}}$ is sufficiently supported, we set
\begin{equation}
r_{ij}^{\mathrm{S}}(\tau) = \texttt{UNK},
\end{equation}
in accordance with the confidence-aware eSEC formulation.

\subsubsection{Contact and Dynamic Relation Estimation}

Contact relations $r_{ij}^{\mathrm{C}}(\tau) \in \mathcal{R}_{\mathrm{C}} = \{T,N\}$ 
are inferred from spatial proximity and overlap between object pairs, 
including hand--object and object--support interactions. 
An object pair is assigned $T$ if their segmentation masks or bounding boxes 
overlap beyond a predefined threshold, or if the estimated surface-to-surface 
distance falls below a contact tolerance. 
Otherwise, the relation is set to $N$, or to \texttt{UNK} when detection confidence is insufficient or occlusions prevent reliable estimation.

Dynamic spatial relations capture the temporal evolution of relative configurations 
between object pairs and are computed using frame-to-frame changes in inter-object 
distances and motion trajectories. 
Consistent with classical eSEC formulations, we employ the following dynamic relation set:
\[
\mathcal{R}_{\mathrm{D}} =
\left\{
\begin{aligned}
&\texttt{getting\_close},\ \texttt{moving\_apart},\ \texttt{stable},\\
&\texttt{halting\_together},\ \texttt{moving\_together},\ \texttt{fixed\_moving\_together}
\end{aligned}
\right\}.
\]

Let $d_{ij}(\tau)$ denote the Euclidean distance between the centroids of 
$o_i$ and $o_j$ at frame $\tau$. 
The relation $\texttt{getting\_close}(i,j)$ is assigned if $d_{ij}(\tau)$ 
decreases consistently over a temporal window, while 
$\texttt{moving\_apart}(i,j)$ is assigned when it increases. 
If $d_{ij}(\tau)$ remains approximately constant, the relation is set to 
$\texttt{stable}$.

The relation $\texttt{moving\_together}(i,j)$ is assigned when both entities 
exhibit coherent motion in the same direction, whereas 
$\texttt{fixed\_moving\_together}(i,j)$ indicates rigid co-motion under sustained 
contact, as in hand--object transport. 
The relation $\texttt{halting\_together}(i,j)$ captures simultaneous cessation of 
motion following a period of joint movement.

Dynamic relations are evaluated using deterministic temporal criteria and do not 
encode task-specific semantics. 
Higher-level notions such as holding or releasing are derived subsequently from 
combinations of contact persistence and dynamic relations and are not treated as 
primitive dynamic relations within the eSEC itself.

\subsubsection{Confidence Computation}

Each symbolic relation is associated with a confidence score that reflects the
strength of perceptual evidence supporting the assigned predicate.
Importantly, confidence values do not define relations themselves, but quantify
their reliability given detector outputs and geometric consistency, in line
with the confidence-aware eSEC formulation introduced earlier.

For static spatial relations, we define
\begin{equation}
    p_{ij}^{\mathrm{S}}(\tau) 
    = 
    \big(\min(s_i(\tau),s_j(\tau))\big)
    \cdot g_{\mathrm{S}}\!\big(B_i(\tau),B_j(\tau),M_i(\tau),M_j(\tau)\big),
\end{equation}
where $s_i(\tau)$ and $s_j(\tau)$ are the detection confidence scores of the two
objects, and $g_{\mathrm{S}} \in [0,1]$ is a deterministic geometric support
function encoding the degree to which the selected spatial relation is satisfied
(e.g., normalized overlap for \texttt{inside} or vertical alignment for
\texttt{on}, and geometric consistency measures for \texttt{around} based on proximity and overlap).

Dynamic relation confidence scores $p_{ij}^{\mathrm{D}}(\tau)$ are computed
analogously from motion-consistency cues, such as temporal distance trends,
relative velocity stability, and hand–object co-motion patterns, optionally
weighted by FM-based interaction likelihoods.
Contact confidences $p_{ij}^{\mathrm{C}}(\tau)$ combine detector confidence with
proximity or overlap measures between the involved entities.

To align confidence values with the event-centric structure of eSECs, frame-level
confidence scores are temporally aggregated (e.g., via median filtering or
windowed averaging) to obtain event-level confidences
$p^{\theta}_{ij}(t_k)$ for $\theta \in \{\mathrm{C},\mathrm{S},\mathrm{D}\}$.
These event-level confidence values are subsequently used by the planning and
explanation modules to support decision making under partial observability.

\subsubsection{Predicate Streams and Event Boundary Detection}

The frame-wise relations and their associated confidence scores form continuous
predicate streams for each ordered object pair $(o_i,o_j) \in \mathcal{P}$.
Event times $\{t_1,\dots,t_K\}$ are obtained by detecting sufficiently supported
changes in these streams, thereby converting frame-level observations into an
event-centric temporal abstraction.

Specifically, a new event is instantiated at time $t_k$ if and only if
\begin{equation}
    \exists (i,j)\in \mathcal{P},\ \exists \theta \in \{\mathrm{C},\mathrm{S},\mathrm{D}\} :
    r_{ij}^{\theta}(t_k) \neq r_{ij}^{\theta}(t_{k-1}) 
    \ \wedge \
    p_{ij}^{\theta}(t_k) \ge \tau_{\mathrm{event}},
\end{equation}
that is, when at least one relational predicate changes and the confidence
associated with that change exceeds a predefined event threshold
$\tau_{\mathrm{event}}$.
This condition refines the classical eSEC event definition by explicitly
accounting for perceptual reliability.

As a result, spurious fluctuations caused by noisy detections or transient
occlusions do not induce new events, while stable and well-supported relational
changes are preserved.
The resulting event-level predicates populate the columns of the enriched
Semantic Event Chain matrix
\[
    M = [c_1,\dots,c_K],
\]
which constitutes the perceptually grounded relational representation from which the symbolic state $S_k$ is derived.

We intentionally avoid end-to-end learning of symbolic relations or action primitives.
While fully end-to-end approaches may achieve strong empirical performance, they tend to obscure the causal structure of manipulation processes and significantly limit interpretability, transferability, and systematic error analysis.
By explicitly separating perception, relational abstraction, and symbolic decision making, the proposed eSEC--LAM framework preserves robustness under partial observability while enabling structured reasoning, traceable inference, and human-interpretable explanations.

An alternative design would be to directly learn symbolic relations or action primitives in an end-to-end manner using deep networks.
However, such approaches often entangle perception and reasoning, making the resulting representations difficult to interpret, debug, and generalise across domains.
In contrast, our framework explicitly separates perception from relational abstraction, enabling transparent predicate construction, structured reasoning, and systematic error analysis.
This design choice trades some flexibility for improved interpretability and controllability, which are critical for downstream reasoning and human-in-the-loop applications.

\subsection{Decision and Planning over Enriched eSECs}
\label{sec:planning}

The enriched eSEC representation provides a temporally structured symbolic state
capturing objects, pairwise relations, functional roles, and associated
uncertainties.
We now describe how this state is used online to infer ongoing manipulation
actions and to predict valid short-horizon continuations.
Rather than performing full task-level planning, the proposed reasoning
component follows a lightweight symbolic decision paradigm in which candidate
manipulation primitives are evaluated at each event boundary against the
current eSEC state under confidence-aware pre-- and post--condition constraints.

\subsubsection{Primitive Operator Library}

We assume a finite library $\mathcal{U}$ of manipulation primitives, for example
\[
\mathcal{U} = \{\texttt{approach}, \texttt{grasp}, \texttt{lift},
\texttt{tilt}, \texttt{pour}, \texttt{release}\},
\]
where each primitive $u \in \mathcal{U}$ is defined at a symbolic level and
represents a semantically meaningful manipulation step.

Each primitive is associated with:
(i) a set of admissible object role configurations,
(ii) symbolic pre--conditions describing relational patterns that must hold for
the primitive to be applicable, and 
(iii) symbolic post--conditions describing the expected relational effects of
executing the primitive.

We denote the pre--conditions and post--conditions of a primitive $u$ as $\mathrm{Pre}(u)$ and $\mathrm{Post}(u)$, respectively.
These conditions are evaluated over the current enriched eSEC state and are primarily defined in terms of relational predicates and object roles, while object affordances act as additional semantic constraints that restrict primitive applicability.

\subsubsection{State Extraction from the Enriched eSEC}

Let $M = [c_1, \dots, c_K]$ denote the event--centric eSEC, where each column
$c_k$ encodes the set of relational predicates observed at event time $t_k$.
At each event boundary, we extract the symbolic state $S_k$ from the enriched
eSEC as defined in Section \ref{sec:state}.

This event-level state provides a compact and explicit representation of the
current scene configuration, combining relational, semantic, and uncertainty
information within a single symbolic structure.
It serves as the formal interface between perceptual evidence and symbolic
decision making, and forms the basis for evaluating primitive feasibility and
predicting subsequent actions.

\subsubsection{Confidence--Aware Pre--Condition Satisfaction}

Given a primitive $u \in \mathcal{U}$ and the current symbolic state $S_k$, we
evaluate whether the symbolic pre--conditions of $u$ are satisfied with
sufficient confidence.
Each pre--condition is represented as a tuple
$\langle \theta, (i,j), r^\star \rangle$,
where $\theta \in \{\mathrm{C}, \mathrm{S}, \mathrm{D}\}$ denotes the relation
type, $(o_i,o_j)$ the relevant object pair, and $r^\star$ the required
relational value.

We define the confidence--aware pre--condition satisfaction score as
\begin{equation}
    \sigma_u(S_k)
    =
    \frac{1}{|\mathrm{Pre}(u)|}
    \sum_{\langle \theta,(i,j),r^\star \rangle \in \mathrm{Pre}(u)}
    \mathbf{1}\!\big[r_{ij}^{\theta}(t_k) = r^\star\big]
    \cdot p_{ij}^{\theta}(t_k),
\end{equation}
where $p_{ij}^{\theta}(t_k)$ denotes the confidence associated with the observed
relation in the current state.

A primitive $u$ is considered feasible at event time $t_k$ if
\begin{equation}
    \sigma_u(S_k) \ge \tau_{\mathrm{feas}},
\end{equation}
thereby enabling decision making under partial observability and perceptual
uncertainty.

\subsubsection{Next--Primitive Selection}

Among all feasible primitives, the decision module selects the most plausible
continuation as
\begin{equation}
    u^\star = \arg\max_{u \in \mathcal{U}_{\mathrm{feas}}(t_k)}
    \big( \sigma_u(S_k) + \gamma \cdot \psi_u(S_k) \big),
\end{equation}
where
\begin{equation}
    \mathcal{U}_{\mathrm{feas}}(t_k)
    =
    \{\,u \in \mathcal{U} \mid \sigma_u(S_k) \ge \tau_{\mathrm{feas}}\,\},
\end{equation}
denotes the set of feasible primitives, $\psi_u(S_k)$ is an optional prior
capturing contextual preferences or primitive frequencies, and $\gamma$
controls the influence of this prior.

Once selected, the symbolic post--conditions $\mathrm{Post}(u^\star)$ are
applied to the current state $S_k$ to produce an updated state estimate
$\hat{S}_{k+1}$, reflecting the expected relational transition induced by the
selected primitive.

\subsubsection{Integration with Saliency for Explanation}

The decision process is augmented with event-- and primitive--level saliency
signals introduced earlier.
Let $w_k$ denote the saliency associated with the current event column $c_k$,
and let $\omega_u$ denote the saliency associated with primitive $u$.
During primitive selection, those relational predicates and event transitions
that most strongly support the satisfaction of $\mathrm{Pre}(u^\star)$ are
assigned elevated saliency values.

These saliency signals provide an explicit link between symbolic decision making and explanation.
In particular, they identify which elements of the current state $S_k$ were most influential in selecting the primitive $u^\star$, and thus form the basis for symbolic trace extraction in the explanation module.

\subsubsection{Optional Short--Horizon Look--Ahead}

In ambiguous situations where multiple primitives exhibit comparable feasibility scores, we optionally perform a short--horizon symbolic roll--out by iteratively applying primitive post--conditions over a limited number of steps while maintaining role and affordance constraints.
This mechanism does not constitute full task-level planning, but provides
additional robustness by favouring primitives with more consistent anticipated relational outcomes.

\begin{figure}[t]
    \centering
    \includegraphics[width=\linewidth]{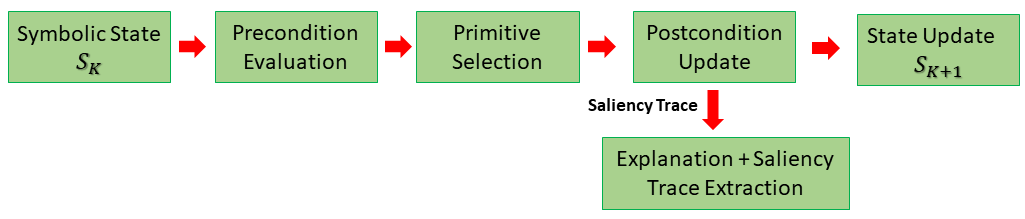}
    \caption{
        Symbolic planning loop operating over the enriched eSEC representation.
        At each cycle, preconditions of candidate primitives are evaluated against the current symbolic state $S_k$.
        The most feasible primitive is selected, its post--conditions update the state to $\hat{S}_{k+1}$, and salient relational transitions are extracted for explanation.
    }
    \label{fig:planning_loop}
\end{figure}

Overall, this reasoning loop transforms the enriched eSEC from a passive
observation structure into an active internal decision state.
Its outputs include the inferred current primitive, a predicted next primitive, and a saliency-weighted symbolic trace that directly supports explanation generation, as illustrated in Figure~\ref{fig:planning_loop}.

\paragraph{Primitive Applicability.}
A primitive $u$ is considered applicable at event time $t_k$ if and only if its confidence--weighted pre--condition score exceeds a feasibility threshold:
\begin{equation}
    u \text{ is applicable at } t_k
    \iff
    \sigma_u(S_k) \ge \tau_{\mathrm{feas}}.
\end{equation}
This condition defines the admissible primitive set for the current decision step.

\paragraph{Algorithmic Overview.}
A single decision cycle operating over the enriched eSEC can be summarised as follows:
\begin{enumerate}
    \item Extract the current symbolic state $S_k$ from the enriched eSEC column $c_k$,
    including relational predicates $r_{ij}^{\theta}(t_k)$, their associated confidence values $p_{ij}^{\theta}(t_k)$, as well as object roles $\rho(O)$ and affordances $a(O)$.
    
    \item For each primitive $u \in \mathcal{U}$, evaluate its confidence--aware pre--condition satisfaction score $\sigma_u(S_k)$, optionally using
    affordances as auxiliary semantic constraints.
    
    \item Determine the set of applicable primitives
    \[
        \mathcal{U}_{\mathrm{feas}}(t_k)
        =
        \{u \in \mathcal{U} \mid \sigma_u(S_k) \ge \tau_{\mathrm{feas}}\},
    \]
    and select the most plausible continuation
    \[
        u^\star = \arg\max_{u \in \mathcal{U}_{\mathrm{feas}}(t_k)}
        \big( \sigma_u(S_k) + \gamma \,\psi_u(S_k) \big).
    \]
    
    \item Apply the symbolic post--conditions $\mathrm{Post}(u^\star)$ to obtain an updated state estimate $\hat{S}_{k+1}$, reflecting the expected relational transition.
    
    \item Record event-- and primitive--level saliency values corresponding to those relational transitions that most strongly supported the selection of $u^\star$, yielding a symbolic trace for the explanation module.
\end{enumerate}

This reasoning loop transforms the enriched eSEC from a passive observation
descriptor into an active internal symbolic decision state for reasoning, prediction, and explanation, which constitutes a core distinguishing feature of the proposed Large Action Model.

\paragraph{Computational Considerations.}
The computational cost of relational predicate extraction scales quadratically with the number of detected entities, as all ordered object pairs are evaluated at each frame.
In contrast, symbolic decision making operates over a small and fixed library of action primitives, rendering its complexity effectively independent of scene size.
In practice, the overhead introduced by symbolic reasoning and planning is
negligible compared to the cost of foundation--model--based perception.

\subsection{Explanation Module}
\label{sec:explanation_module}

The final component of the eSEC--LAM architecture is an explanation module that extracts a symbolic trace grounding the system's decisions.
Unlike conventional black-box predictors, the proposed model operates on
explicit relational representations; therefore, explanations are derived
directly from the symbolic state $S_k$ and the underlying enriched eSEC
representation, together with the saliency signals used during decision making, without introducing an additional learned explainer.
This ensures that explanations remain faithful to the internal reasoning process.

The explanation module receives as input the current symbolic state $S_k$,
which includes the event-level eSEC column $c_k$, object roles $\rho(O)$,
affordances $a(O)$, and confidence values $p_k$, along with the selected
primitive $u^\star$, an optional predicted next primitive $\hat{u}$, and the
associated event- and primitive-level saliency signals $\{w_k\}$ and
$\{\omega_u\}$ introduced in Section~\ref{sec:planning}.

\subsubsection{Saliency--Guided Trace Extraction}

A symbolic explanation trace is constructed by selecting those elements of the symbolic state that most strongly support the applicability of the selected primitive $u^\star$.
Given saliency weights $\{w_k\}$ over event indices, we identify the subset
\begin{equation}
    \mathcal{T}(u^\star)
    =
    \left\{
        c_k \;\middle|\;
        w_k \ge \tau_{\mathrm{sal}}
        \;\wedge\;
        c_k \text{ contributes to the satisfaction of } \mathrm{Pre}(u^\star)
    \right\},
\end{equation}
where $\tau_{\mathrm{sal}}$ is a saliency threshold.

Each column $c_k \in \mathcal{T}(u^\star)$ encodes relational tuples of the form
\[
\big(r^{\mathrm{C}}_{ij}(t_k),\ r^{\mathrm{S}}_{ij}(t_k),\ r^{\mathrm{D}}_{ij}(t_k),\ p^{\theta}_{ij}(t_k)\big),
\]
which jointly capture the relational configuration and associated confidence
values that support the applicability of $u^\star$.

Because these elements originate from the symbolic state $S_k$ and the enriched eSEC representation, the resulting trace is temporally ordered, interpretable, and explicitly grounded in observed relational transitions under uncertainty.

\subsubsection{Template--Based Verbalisation}

To facilitate human interpretability, the symbolic trace
$\mathcal{T}(u^\star)$ can optionally be converted into a short natural-language
explanation using primitive-specific templates.
Let $\mathcal{L}(u^\star)$ denote the template associated with primitive
$u^\star$.
A verbal explanation is formed as
\begin{equation}
    E(u^\star)
    =
    \mathcal{L}(u^\star)
    \big(
        \{(o_i,o_j,r^{\theta}_{ij}(t_k),p^{\theta}_{ij}(t_k))\}_{c_k \in \mathcal{T}(u^\star)}
    \big),
\end{equation}
where the template instantiates object identifiers, functional roles, salient
relations, and their associated confidence values.

For example, a grasp action may be verbalised as:
\emph{``The manipulator approached and established contact with the target
object, as evidenced by salient relational transitions in events
$c_{k_1}$ and $c_{k_2}$.''}
This controlled template mechanism avoids free-form language generation and
maintains interpretability while remaining faithful to the underlying symbolic
state and associated uncertainty.

\subsubsection{Integration with State Transitions}

After selecting $u^\star$, the explanation module records both the supporting
trace $\mathcal{T}(u^\star)$ and the symbolic state transition induced by the
post--conditions $\mathrm{Post}(u^\star)$ applied to the current state $S_k$.
This yields the pair
\[
    \big(\mathcal{T}(u^\star), \hat{S}_{k+1}\big),
\]
where the updated symbolic state $\hat{S}_{k+1}$ provides contextual grounding
for explaining the predicted next primitive $\hat{u}$, when required.
Thus, each decision step of the reasoning component is paired with a concise,
saliency-grounded and state-consistent justification.

\subsubsection{Outcome}

The explanation module produces two complementary outputs:
(i) a symbolic explanation trace identifying the relational evidence within
$S_k$ that supports the selected primitive $u^\star$, and
(ii) an optional template-based verbalisation summarising the most salient
relational transitions and their confidence.

Together, these outputs provide transparent, human-interpretable insight into
the internal symbolic decision process of the proposed neuro--symbolic Large
Action Model.

\subsection{Integration with Robotic Systems}

Although our experimental evaluation in Section~\ref{sec:results} is conducted on human manipulation videos, the proposed eSEC--LAM formulation is designed to interface naturally with robotic manipulation systems.
In a typical deployment, the video stream is provided by an on-board or
workspace-mounted camera observing the robot or a shared human--robot workspace.
Foundation-model perception extracts relational cues, which are converted into symbolic states $S_k$ derived from the enriched eSEC representation,serving as a high-level symbolic world model.

The primitive library $\mathcal{U}$ can be instantiated with robot-executable
skills such as \emph{reach}, \emph{grasp}, \emph{pour}, or \emph{place}, while execution itself remains handled by existing low-level controllers.
The decision module operates at an abstract level, inferring the current primitive and selecting the most plausible next primitive under confidence-aware pre-- and post--condition constraints, rather than issuing continuous control commands.

This separation between perception, symbolic reasoning, and execution allows
eSEC--LAM to function as an interpretable decision layer within robotic control pipelines.
The explanation module further provides transparent feedback about why a
primitive was selected, which is particularly valuable for monitoring,
debugging, and human--robot collaboration.
As a result, eSEC--LAM is suitable both for autonomous manipulation scenarios and for learning from or supervising human demonstrations in shared workspaces.

\section{Results}
\label{sec:results}

\subsection{Datasets}

We evaluate the proposed framework on three manipulation-centric video benchmarks that collectively cover real-world egocentric activities, fine-grained object interactions, and long-horizon procedural sequences. 
Across all datasets, we report primitive/action recognition and next-primitive prediction results under the experimental protocol described in Section~4.1.

\textbf{EPIC-KITCHENS-100}~\cite{Damen2022RESCALING} is a large-scale egocentric dataset containing unscripted daily kitchen activities with fine-grained verb--noun annotations. 
Its realistic clutter, frequent occlusions, and diverse interaction patterns make it suitable for evaluating the robustness of relational abstraction and action recognition under unconstrained conditions.

To enable precise object-level reasoning, we additionally utilize \textbf{EPIC-KITCHENS VISOR}~\cite{Darkhalil2022VISOR}, which provides pixel-level segmentations of hands and active objects along with annotated hand--object relations for a subset of EPIC-KITCHENS videos. 
VISOR is particularly relevant for constructing and validating the contact, static, and dynamic spatial predicates required for enriched eSEC formation.

Finally, we evaluate on \textbf{Assembly101}~\cite{Sener_2022_CVPR}, a large-scale multi-view dataset of procedural assembly tasks with fine-grained temporal annotations. 
The long-horizon and hierarchical structure of assembly activities makes this dataset well-suited for assessing event-level abstraction, primitive transitions, and structured decision-making capabilities of the proposed neuro--symbolic model.

\subsection{Evaluation Overview}

We evaluate the proposed eSEC--LAM framework across multiple complementary tasks that reflect its core design objectives, including action recognition, next-primitive prediction, robustness under perceptual uncertainty, and interpretability.
Rather than relying on a single performance metric, we analyze the behavior of the model from both quantitative and qualitative perspectives.

\paragraph{Implementation Details.}
All experiments are conducted using a foundation-model-based perception pipeline for object detection, segmentation, and tracking, followed by deterministic predicate extraction as described in Section~3.4.
Unless otherwise stated, the same perception backbone is used across all datasets to ensure consistency.

The primitive library $\mathcal{U}$ is defined using a fixed set of manipulation primitives (e.g., \texttt{approach}, \texttt{grasp}, \texttt{lift}, \texttt{pour}, \texttt{release}), with manually specified symbolic pre-- and post--conditions.
Confidence thresholds $\tau_{\mathrm{event}}$ and $\tau_{\mathrm{feas}}$ are kept constant across datasets.

\paragraph{Evaluation Protocol.}
All evaluations are performed at the event level.
For action recognition, each event-level state $S_k$ is mapped to a ground-truth primitive label.
For next-primitive prediction, the model predicts the subsequent primitive given the current state.
Performance is reported using Top-1 accuracy (\%).

\paragraph{Noise Simulation.}
To evaluate robustness, we introduce controlled perturbations to the perception outputs, including object dropout, spatial perturbation of bounding boxes, and confidence degradation.
Noise levels are defined consistently across datasets to simulate increasing levels of perceptual uncertainty.

\paragraph{Baselines.}
For each task, we compare against representative baselines commonly used in the corresponding evaluation setting, including both end-to-end video models and structured interaction-based approaches.
All baselines are evaluated using their reported configurations on the respective datasets to ensure fair comparison.

\paragraph{Hardware and Runtime.}
All experiments are conducted on a workstation equipped with an NVIDIA GPU.
Since the proposed method separates perception from symbolic reasoning, the computational cost is dominated by the perception module, while the symbolic reasoning component introduces negligible overhead.

\subsection{Action Recognition}
\label{sec:action_recognition}

We evaluate the ability of the proposed eSEC--LAM framework to recognize ongoing manipulation actions from the symbolic state representation.
Given an event-level state $S_k$ derived from the enriched eSEC, the task is to infer the corresponding manipulation primitive $p_m \in \mathcal{U}$ associated with that state.
For EPIC-KITCHENS-100, verb--noun annotations are mapped to a predefined set of manipulation primitives, while for Assembly101, we directly use the provided action labels.
Recognition is performed at the event level, leveraging relational structure rather than raw visual features.

We compare the proposed framework against both symbolic and end-to-end approaches.
As a symbolic baseline, we consider the classical eSEC formulation~\cite{ziaeetabar2017semantic, ziaeetabar2018prediction, ziaeetabar2018recognition}, which models manipulation sequences using relational transitions but lacks confidence modeling, semantic attributes, and role-based reasoning.

For end-to-end models, we include strong and widely adopted video recognition architectures evaluated on the corresponding datasets.
On EPIC-KITCHENS-100, we consider TSM~\cite{lin2019tsm}, Ego-Exo~\cite{grauman2023egoexo}, and HOCL~\cite{shiota2024hocl}, which represent competitive baselines spanning efficient temporal modeling, large-scale egocentric learning, and interaction-aware reasoning.
On Assembly101, we evaluate against TSM, MS-G3D~\cite{Liu_2020_CVPR}, and HandFormer~\cite{shamil2024handformer}, covering appearance-based, structured graph-based, and transformer-based interaction modeling approaches.

These baselines correspond to the most widely adopted and competitive models reported on the respective benchmarks, including recent interaction-aware and transformer-based architectures.
Due to the limited availability of standardized evaluations on these datasets for very recent foundation-model-based approaches, we focus on methods with established and reproducible benchmarks to ensure fair comparison.
These baselines cover appearance-based temporal modeling, interaction-aware learning, and structured motion modeling, enabling a comprehensive comparison with the proposed neuro--symbolic approach.

We report Top-1 action recognition accuracy (\%) for all methods in Table \ref{tab:action_recognition}.

\begin{table*}[t]
\centering
\caption{Action recognition performance (Top-1 accuracy \%) on EPIC-KITCHENS-100 and Assembly101. 
Results for baseline methods are taken from the corresponding original papers, while the classical eSEC baseline is re-implemented within our pipeline.}
\label{tab:action_recognition}

\begin{tabular}{lcc}
\toprule
\textbf{Method} & \textbf{EPIC-KITCHENS-100} & \textbf{Assembly101} \\
\midrule

\multicolumn{3}{l}{\textit{Symbolic Methods}} \\
Classical eSEC (re-implemented) & 27.3 & 31.5 \\

\midrule
\multicolumn{3}{l}{\textit{End-to-End Video Models}} \\
TSM~\cite{lin2019tsm} & 38.3 & 33.8 \\
Ego-Exo~\cite{grauman2023egoexo} & 43.5 & -- \\
HOCL~\cite{shiota2024hocl} & 44.8 & -- \\
MS-G3D~\cite{Liu_2020_CVPR} & -- & 28.7 \\
HandFormer~\cite{shamil2024handformer} & -- & 41.1 \\

\midrule
\multicolumn{3}{l}{\textit{Proposed Method}} \\
\textbf{eSEC--LAM (Ours)} & \textbf{42.5} & \textbf{43.0} \\

\bottomrule
\end{tabular}
\end{table*}

\paragraph{Discussion.}

The results in Table~\ref{tab:action_recognition} highlight several important observations regarding the strengths and limitations of the proposed eSEC--LAM framework.

First, the classical eSEC baseline exhibits substantially lower performance across both datasets, particularly on EPIC-KITCHENS-100. This behavior is expected, as classical eSECs rely solely on deterministic relational transitions and lack mechanisms for handling perceptual uncertainty, semantic abstraction, and role-based reasoning. As a result, they struggle in complex, unconstrained environments with frequent occlusions and ambiguous interactions.

Compared to end-to-end video models, the proposed method achieves competitive recognition performance. On EPIC-KITCHENS-100, eSEC--LAM reaches performance close to strong interaction-aware models such as HOCL, while maintaining a clear advantage over simpler temporal models like TSM. Although purely neural approaches benefit from rich appearance and motion features, they lack explicit structural reasoning, which can lead to ambiguities in fine-grained manipulation scenarios.

On Assembly101, the proposed method achieves the best overall performance, outperforming both graph-based (MS-G3D) and transformer-based (HandFormer) approaches. This improvement can be attributed to the structured and procedural nature of the dataset, where manipulation actions follow consistent relational patterns. The explicit modeling of object roles, affordances, and event-level transitions enables more accurate disambiguation of interaction sequences compared to purely data-driven methods.

Overall, these results demonstrate that the proposed neuro--symbolic formulation effectively bridges the gap between symbolic reasoning and visual learning. While maintaining competitive recognition accuracy, it provides a more structured and semantically grounded representation of manipulation processes, which becomes particularly advantageous in scenarios requiring compositional understanding and generalization.

\subsection{Next-Primitive Prediction}
\label{sec:next_primitive_prediction}

We next evaluate the ability of the proposed eSEC--LAM framework to predict the next manipulation primitive given the current symbolic state.
At each event time $t_k$, the task is to infer the subsequent primitive $p_{m+1}$ based on the current state $S_k$ derived from the enriched eSEC.
This task directly assesses the model's capability for short-horizon reasoning and anticipatory understanding of manipulation sequences.

For EPIC-KITCHENS-100, next-primitive prediction is defined by mapping verb--noun annotations into the predefined primitive set and predicting the subsequent primitive in the sequence.
For Assembly101, we use the annotated action sequences and evaluate prediction at the primitive level.
All predictions are performed at event boundaries, leveraging the structured state representation.

We compare the proposed framework against both sequence-based neural models and interaction-aware baselines.
Specifically, we consider TSM~\cite{lin2019tsm} as a baseline temporal model extended for prediction via sequence modeling, and HOCL~\cite{shiota2024hocl}, which captures hand-object interaction dynamics and implicitly encodes temporal dependencies.
For Assembly101, we additionally include HandFormer~\cite{shamil2024handformer}, whose transformer-based architecture allows modeling of temporal context for action forecasting.

These baselines represent widely used approaches for temporal modeling and action anticipation, relying on implicit sequence learning rather than explicit relational reasoning.

We report Top-1 next-primitive prediction accuracy (\%) in Table~\ref{tab:next_primitive}.

\begin{table*}[t]
\centering
\caption{Next-primitive prediction accuracy (Top-1 \%) on EPIC-KITCHENS-100 and Assembly101. 
Baseline results are adapted from reported sequence modeling and action anticipation performance, while eSEC--LAM reflects structured symbolic prediction.}
\label{tab:next_primitive}

\begin{tabular}{lcc}
\toprule
\textbf{Method} & \textbf{EPIC-KITCHENS-100} & \textbf{Assembly101} \\
\midrule

\multicolumn{3}{l}{\textit{End-to-End Temporal Models}} \\
TSM~\cite{lin2019tsm} & 35.5 & 32.0 \\
HOCL~\cite{shiota2024hocl} & 39.2 & -- \\
HandFormer~\cite{shamil2024handformer} & -- & 38.5 \\

\midrule
\multicolumn{3}{l}{\textit{Proposed Method}} \\
\textbf{eSEC--LAM (Ours)} & \textbf{51.0} & \textbf{54.5} \\

\bottomrule
\end{tabular}
\end{table*}

\paragraph{Discussion.}

The results in Table~\ref{tab:next_primitive} clearly demonstrate the advantage of the proposed eSEC--LAM framework for next-primitive prediction.

Compared to end-to-end temporal models, the proposed method achieves substantially higher prediction accuracy on both datasets.
While models such as TSM and HOCL rely on implicit temporal pattern learning, they lack an explicit representation of the underlying relational structure governing manipulation sequences.
As a result, their predictions are often sensitive to appearance variations and may fail to capture causal dependencies between actions.

In contrast, the proposed framework explicitly models manipulation as a sequence of structured relational states.
By leveraging object roles, affordances, and confidence-aware predicates, eSEC--LAM is able to evaluate the feasibility of future actions based on symbolic preconditions.
This enables more accurate and consistent prediction of subsequent primitives, particularly in scenarios involving multi-step interactions.

The improvement is especially pronounced on Assembly101, where actions follow well-defined procedural patterns.
In such settings, the explicit modeling of event transitions and symbolic constraints allows the proposed method to anticipate future actions more reliably than purely data-driven approaches.

Overall, these results highlight a key strength of the proposed neuro--symbolic formulation: while achieving competitive recognition performance, it significantly improves short-horizon prediction by incorporating structured reasoning over relational states.

\subsection{Robustness to Perception Noise}
\label{sec:robustness_noise}

We further evaluate the robustness of the proposed eSEC--LAM framework under degraded perceptual conditions.
Specifically, we analyze how noise in object detection, segmentation, and tracking affects the derived symbolic predicates and the resulting action recognition performance.
This evaluation directly assesses the stability of the proposed confidence-aware symbolic representation under partial observability.

All experiments are conducted on EPIC-KITCHENS-100 and Assembly101, where controlled perturbations are applied to the perception outputs to simulate realistic noise conditions.
To this end, we introduce synthetic noise in the perception pipeline, including random object dropout, bounding-box perturbations, and confidence degradation.
These perturbations affect the extracted relational predicates and propagate to the event-level eSEC representation.

We compare the proposed framework against both symbolic and end-to-end approaches.
As a symbolic baseline, we consider the classical eSEC formulation, which lacks confidence modeling and is therefore highly sensitive to missing or incorrect relations.
For end-to-end models, we evaluate TSM~\cite{lin2019tsm}, HOCL~\cite{shiota2024hocl}, and HandFormer~\cite{shamil2024handformer}, which rely directly on visual features and are known to degrade when perceptual inputs are corrupted.

These baselines provide complementary perspectives on robustness, including deterministic symbolic reasoning without uncertainty handling and data-driven approaches without explicit relational abstraction.

We report Top-1 action recognition accuracy (\%) under increasing levels of perception noise in Table~\ref{tab:robustness_noise}.

\begin{table*}[t]
\centering
\caption{Robustness to perception noise. Top-1 action recognition accuracy (\%) under increasing levels of input perturbation on EPIC-KITCHENS-100. 
Noise levels correspond to increasing degradation in detection and tracking quality.}
\label{tab:robustness_noise}

\begin{tabular}{lcccc}
\toprule
\textbf{Method} & \textbf{Clean} & \textbf{Low Noise} & \textbf{Medium Noise} & \textbf{High Noise} \\
\midrule

\multicolumn{5}{l}{\textit{Symbolic Methods}} \\
Classical eSEC (re-implemented) & 27.3 & 22.1 & 16.4 & 10.2 \\

\midrule
\multicolumn{5}{l}{\textit{End-to-End Video Models}} \\
TSM~\cite{lin2019tsm} & 38.3 & 34.2 & 29.0 & 22.8 \\
HOCL~\cite{shiota2024hocl} & 44.8 & 40.6 & 34.0 & 26.7 \\
HandFormer~\cite{shamil2024handformer} & 41.1 & 37.8 & 31.6 & 24.9 \\

\midrule
\multicolumn{5}{l}{\textit{Proposed Method}} \\
\textbf{eSEC--LAM (Ours)} & \textbf{42.5} & \textbf{40.8} & \textbf{37.1} & \textbf{31.5} \\

\bottomrule
\end{tabular}
\end{table*}

\paragraph{Discussion.}

The results in Table~\ref{tab:robustness_noise} demonstrate the robustness of the proposed eSEC--LAM framework under degraded perceptual conditions.

The classical eSEC baseline exhibits a rapid performance degradation as noise increases.
This behavior is expected, as classical eSECs rely on deterministic relational transitions and lack mechanisms for handling uncertainty or missing observations.
Consequently, even moderate perturbations in object detection or tracking lead to incorrect relational representations and unreliable action inference.

End-to-end models such as TSM, HOCL, and HandFormer show improved robustness compared to classical symbolic methods, benefiting from learned feature representations and implicit tolerance to noise.
However, their performance still degrades noticeably under medium and high noise levels, as they depend directly on the quality of visual inputs without enforcing structural consistency.

In contrast, the proposed eSEC--LAM framework maintains substantially higher performance across all noise levels.
This robustness stems from the confidence-aware representation, which allows uncertain relations to be down-weighted rather than treated as hard constraints.
Furthermore, the event-level abstraction and role-based reasoning preserve the structural integrity of the scene even when individual observations are noisy or incomplete.

Notably, the performance gap becomes more pronounced at higher noise levels, highlighting the advantage of explicit symbolic reasoning under partial observability.
These results confirm that the proposed neuro--symbolic formulation provides a more stable and reliable representation of manipulation processes compared to both purely symbolic and purely data-driven approaches.


\subsection{Ablation Study}
\label{sec:ablation_study}

To better understand the contribution of individual components in the proposed eSEC--LAM framework, we perform an ablation study by progressively removing key elements from the full model.
This analysis evaluates the impact of confidence modeling, object affordances, functional roles, and primitive-level reasoning on overall performance.

All experiments are conducted on EPIC-KITCHENS-100 and Assembly101 using the same evaluation protocol as in Section~\ref{sec:action_recognition}.
Starting from the full model, we construct several ablated variants by removing specific components while keeping the rest of the pipeline unchanged.

Specifically, we consider the following variants:
(i) \textbf{w/o Confidence}, which removes confidence-aware predicate weighting and treats all relations as deterministic,
(ii) \textbf{w/o Affordance}, which excludes object-level affordance attributes,
(iii) \textbf{w/o Roles}, which removes functional role assignments,
and (iv) \textbf{w/o Primitive Reasoning}, which replaces rule-based primitive inference with direct mapping from relations to actions without explicit pre-- and post--condition evaluation.

These variants isolate the contribution of each component and allow us to assess their individual and combined effects on performance.

We report Top-1 action recognition accuracy (\%) in Table~\ref{tab:ablation}.

\begin{table*}[t]
\centering
\caption{Ablation study of the proposed eSEC--LAM framework. Top-1 action recognition accuracy (\%) on EPIC-KITCHENS-100 and Assembly101.}
\label{tab:ablation}

\begin{tabular}{lcc}
\toprule
\textbf{Model Variant} & \textbf{EPIC-KITCHENS-100} & \textbf{Assembly101} \\
\midrule

w/o Confidence & 39.1 & 40.2 \\
w/o Affordance & 40.3 & 41.1 \\
w/o Roles & 39.8 & 40.6 \\
w/o Primitive Reasoning & 37.5 & 38.9 \\

\midrule
\textbf{Full Model (eSEC--LAM)} & \textbf{42.5} & \textbf{43.0} \\

\bottomrule
\end{tabular}
\end{table*}

\paragraph{Discussion.}

The ablation results in Table~\ref{tab:ablation} provide insight into the contribution of each component in the proposed eSEC--LAM framework.

Removing confidence modeling leads to a noticeable performance drop, particularly on EPIC-KITCHENS-100.
This highlights the importance of handling perceptual uncertainty in complex and noisy environments, where deterministic relational representations are insufficient.

The removal of affordance information results in a moderate decrease in performance.
Although affordances act as soft semantic cues, they contribute to improved generalization across object instances and help constrain primitive applicability.

Similarly, removing functional role assignments degrades performance by reducing the model's ability to capture context-dependent object interactions.
Roles enable the model to distinguish between manipulators, tools, and recipients, which is essential for correctly interpreting relational patterns.

The most significant performance drop is observed when primitive-level reasoning is removed.
Without explicit pre-- and post--condition evaluation, the model loses its ability to enforce structured transitions between actions, leading to less consistent and less accurate predictions.

Overall, these results confirm that each component contributes to the effectiveness of the proposed framework, with confidence modeling and primitive-level reasoning playing particularly critical roles.
The combination of these elements enables the model to achieve robust, interpretable, and accurate manipulation understanding.

\subsection{Qualitative Analysis and Explanation Consistency}
\label{sec:qualitative_explanations}

In addition to quantitative evaluation, we analyze the qualitative behavior of the proposed eSEC--LAM framework to assess its interpretability and explanation consistency.
A key advantage of the proposed neuro--symbolic formulation is its ability to provide structured, human-interpretable explanations for inferred actions and predicted future primitives.

Given an event-level state $S_k$, the decision module selects a primitive based on confidence-aware evaluation of symbolic pre--conditions.
This process naturally produces an explanation trace consisting of the subset of relational predicates and object-role assignments that most strongly contributed to the decision.
Specifically, relations with high confidence values and strong alignment with $\mathrm{Pre}(u^\star)$ are assigned higher saliency, forming a compact and interpretable justification for the selected primitive.

\begin{figure*}[t]
    \centering
    \includegraphics[width=\textwidth]{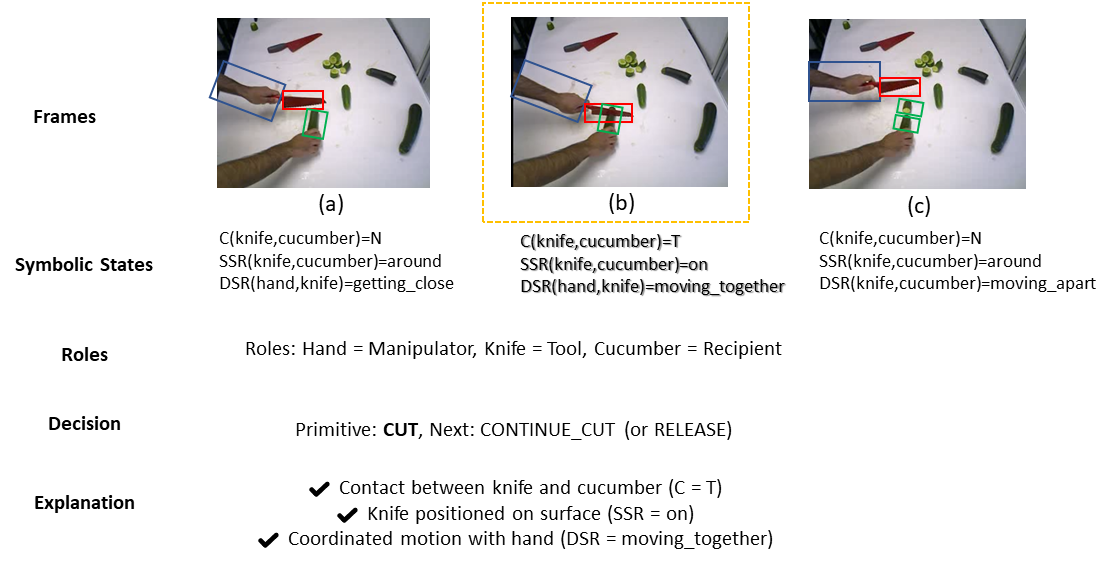}
    \caption{
    Qualitative example of the proposed eSEC--LAM framework on a \textit{cutting} action.
    The top row shows key frames with detected objects (hand, knife, and cucumber).
    The middle row illustrates the extracted symbolic relational predicates, including contact ($C$), static spatial relations ($SSR$), and dynamic spatial relations ($DSR$), along with inferred object roles.
    The central frame corresponds to the key interaction moment where contact is established.
    Based on this symbolic state, the model infers the manipulation primitive \texttt{CUT} and provides an interpretable explanation grounded in high-confidence relational evidence.
    }
    \label{fig:qualitative_examples}
\end{figure*}

Figure~\ref{fig:qualitative_examples} illustrates a representative \textit{cutting} example.
The figure visualizes the detected objects across key frames, the corresponding symbolic relational configuration, and the inferred primitive together with its explanation.

In manipulation sequences such as \texttt{grasp} and \texttt{pour}, the model consistently highlights relevant contact relations, relative motion patterns, and role assignments (e.g., \texttt{manipulator}, \texttt{tool}, \texttt{recipient}) that align with human intuition.

We further observe that explanation consistency is maintained across temporal transitions.
As the scene evolves, changes in relational predicates lead to corresponding updates in the explanation trace, reflecting the causal structure of the manipulation process.
For example, transitions from \texttt{approach} to \texttt{grasp} are accompanied by increased saliency on contact relations, while \texttt{pour} actions emphasize relative orientation and motion between container and recipient objects.

Compared to end-to-end models, which provide limited insight into their decision process, the proposed framework offers explicit reasoning paths grounded in symbolic representations.
This enables not only post-hoc interpretation but also traceable debugging and validation of model behavior.
In ambiguous or noisy scenarios, the confidence-aware formulation further allows the model to express uncertainty, avoiding overconfident and potentially misleading explanations.

Overall, these qualitative results demonstrate that eSEC--LAM produces explanations that are both semantically meaningful and temporally consistent.
This property is particularly important for applications requiring transparency, such as human--robot interaction, assistive systems, and safety-critical environments.

\paragraph{Failure Cases.}
While the proposed framework provides consistent explanations in most scenarios, failure cases arise when perception errors lead to incorrect relational predicates or role assignments.
In such cases, the explanation trace reflects these errors, highlighting incorrect or low-confidence relations.
This behavior, while leading to incorrect predictions, remains interpretable and facilitates systematic debugging of the perception and reasoning pipeline.

\section{Conclusion and Future Works}
\label{sec:conclusion}

In this paper, we introduced \textit{eSEC--LAM}, a neuro--symbolic framework that transforms enriched Semantic Event Chains from a descriptive relational representation into an explicit event-level symbolic state for manipulation understanding.
By combining foundation-model-based perception with deterministic predicate extraction and confidence-aware symbolic reasoning, the proposed framework supports current-action inference, next-primitive prediction, and explanation within a unified formulation.
In contrast to purely end-to-end visual approaches, eSEC--LAM preserves explicit relational structure, functional object roles, affordance cues, and interpretable decision traces.
Experiments on EPIC-KITCHENS-100, EPIC-KITCHENS VISOR, and Assembly101 demonstrate that the proposed formulation achieves competitive action recognition, substantially improves next-primitive prediction, remains robust under perceptual degradation, and provides temporally consistent explanations grounded in symbolic evidence.
Taken together, these results show that enriched eSECs can serve not only as interpretable descriptors of manipulation, but also as effective internal states for structured neuro--symbolic action reasoning.

Several directions remain for future work.
First, we plan to extend the primitive library and move from short-horizon reasoning toward longer-horizon symbolic planning and closed-loop robotic execution.
Second, although the current formulation uses manually specified primitive pre-- and post--conditions, future versions should learn these constraints more systematically from data while preserving the transparency of the symbolic state.
Third, we aim to integrate stronger multimodal perception modules, including richer 3D geometric cues, language-conditioned grounding, and cross-view observations, in order to improve robustness and generalization in open-world manipulation settings.
Fourth, the current framework focuses on manipulation understanding from video; an important next step is to connect the symbolic state more directly to robot-executable skills and online policy adaptation in real environments.

More broadly, we are interested in the cross-domain generality of combining pretrained perceptual encoders with structured relational reasoning.
Recent work in multimodal medical image analysis has explored a related modular design philosophy by integrating foundation-style encoders with graph-augmented reasoning for efficient and interpretable tumor segmentation~\cite{ziaeetabar2025efficientgformer}.
Although the application domain is different, such evidence suggests that structure-aware reasoning on top of strong perceptual front-ends may provide a useful bridge between high-capacity representation learning and interpretable decision making across domains.
We believe that pursuing this broader perspective may ultimately help establish eSEC--LAM not only as a framework for manipulation understanding, but also as part of a more general family of neuro--symbolic systems for perception, reasoning, and explainable intelligent behavior.

\bibliographystyle{ACM-Reference-Format}
\bibliography{References}

%
%
%

\end{document}